\documentclass[11pt]{article}

\usepackage[preprint]{acl}

\usepackage{times}
\usepackage{latexsym}

\usepackage[T1]{fontenc}
\usepackage[utf8]{inputenc}
\usepackage{microtype}
\usepackage{inconsolata}

\usepackage{graphicx}
\usepackage{booktabs}
\usepackage{amsmath}
\usepackage{amssymb}
\usepackage{bm}
\usepackage{cleveref}
\usepackage{multirow}
\usepackage{multicol}
\usepackage[most]{tcolorbox}
\tcbuselibrary{breakable}
\usepackage{makecell}

\newtcolorbox{promptbox}[1]{
  colback=gray!5,
  colframe=gray!60,
  title={#1},
  fonttitle=\bfseries,
  boxrule=0.5pt,
  arc=2pt,
  left=6pt,
  right=6pt,
  top=6pt,
  bottom=6pt,
}

\renewcommand{\thefootnote}{\fnsymbol{footnote}}

\title{Human-Anchored Factuality Evaluation with Strategic Annotation}

\author{Yu Wang \\
  Amazon AGI \\
  \texttt{ywyu@amazon.com} \\\And
  Craig Erickson \\
  Amazon AGI \\
  \texttt{craigeri@amazon.com} \\\And
  Kevin Small$^\dagger$ \\
  Amazon AGI \\
  \texttt{kevinsmall@gmail.com}
  \\}

\begin{document}
\maketitle

\footnotetext[2]{Now at Microsoft.}
\renewcommand{\thefootnote}{\arabic{footnote}}
\setcounter{footnote}{0}

\begin{abstract}
LLM-based factuality judges provide scalable evaluation signals, but their metrics are often systematically biased relative to human judgments. We study human-anchored factuality evaluation under limited annotation budgets, where judge predictions on the full dataset are combined with human labels on a small selectively sampled subset to obtain statistically valid estimates. The efficiency of this approach depends critically on which examples receive human annotation: in factuality evaluation, judge-human misalignment is not driven solely by low confidence, but also by structured failure modes such as incomplete evidence, temporal mismatch, unverifiable claims, and rubric misalignment. To exploit this structure, we introduce a factuality-specific annotation policy design pipeline that uses failure-space analysis (FSA) to derive diverse predictive signals for modeling human-judge misalignment. On an internal reference-based factuality evaluation system (AutoFA) and RAGTruth, where judge-predicted estimates substantially underestimate human-annotated factual accuracy, our FSA-guided policy improves annotation efficiency over uniform sampling and uncertainty-driven baselines, achieving effective-sample-size gains of 40.3\% on AutoFA and 27.1\% on RAGTruth.
\end{abstract}

\section{Introduction}
\label{sec:introduction}

Factuality evaluation is a recurring concern when deploying large language models in user-facing systems. In production settings, teams rely on factual accuracy to measure system quality (e.g., LLM model, prompts, grounding retrieval), monitor regressions, and make release decisions from evaluation data that are both timely and statistically reliable. Human annotation remains the most trusted way to measure factual accuracy, but it is expensive and slow. Collecting sufficient labels to obtain narrow confidence intervals requires substantial annotation budget and introduces operational latency. This creates a practical tension between evaluation quality and evaluation efficiency.

LLM-as-a-judge (LLMaaJ) systems offer an attractive way to scale factuality evaluation \citep{factscore, felm, ragtruth, safe, veriscore}. Given a query, a response, and reference evidence, the judge can decompose the response into claims, verify them against evidence, and produce factuality labels at much larger scale than human annotators. However, judge-predicted factuality estimates are not necessarily human-anchored \citep{mtbench, llmnotfair}. In reference-based factuality evaluation, judge errors are often systematic rather than random. A judge may reject reasonable answers when references are incomplete, stale, conflicting, or indirectly supportive. It may also apply a more literal verification rubric than human annotators. As a result, treating judge predictions as ground truth can produce biased factual accuracy estimates, even when the judge is useful as a proxy signal.

This paper studies factuality evaluation under a budgeted annotation setting. Given a large set of evaluation items, an LLMaaJ provides proxy factuality predictions for all items, while only a small subset can receive human labels. The goal is to estimate the human-defined factual accuracy rate (FAR) with valid uncertainty quantification and lower variance than uniformly sampled human annotation alone. We adopt Active Statistical Inference (ASI)~\citep{asi}, which combines judge predictions with human corrections from a sampled subset of examples. ASI preserves the human-anchored target, but its efficiency depends critically on the annotation policy -- human labels should be allocated to examples where the judge is most likely to disagree with humans.

Factuality residuals are often structured rather than uniformly distributed across examples. Existing confidence-based approaches~\citep{cdi} use the judge's expressed uncertainty as a sampling signal, but judge-human misalignment is not driven solely by low confidence. A judge can be confidently misaligned when evidence is incomplete, temporally inconsistent, or only indirectly supportive, and when responses require paraphrase, inference, or rubric tolerance that humans handle differently from the judge. To exploit this structure, we construct the annotation policy using \emph{failure-space analysis} (FSA), a factuality-specific policy-design pipeline that derives predictive signals from judge-output, evidence quality, task and input strata, and answer-rubric alignment before converting them into ASI sampling policies.

We evaluate the framework on two benchmarks: an internal Automatic Factual Evaluation system (AutoFA) for virtual assistant responses, and RAGTruth, a public hallucination benchmark with human span-level annotations. In both settings, judge-predicted FAR substantially underestimates the human-annotated FAR and adopting FSA-derived policies preserve desired coverage while improving annotation efficiency over uniform sampling and confidence-based sampling. Aggregated across budgets, FSA-guided ASI achieves an average effective sample size (ESS) gain of 40.3\% on AutoFA and 27.1\% on RAGTruth.

To summarize, our contributions are threefold. First, we formulate reference-based factuality evaluation as a human-anchored, judge-assisted inference problem under limited annotation budget. Second, we introduce failure-space analysis as a practical policy-design layer for identifying factuality-specific judge-human residual risk. Third, we demonstrate on both internal and public factuality settings that FSA-guided ASI recovers the human-defined evaluation target while improving annotation efficiency over uniform and confidence-based alternatives. 
Together, these contributions enable reliable and cost-efficient factuality measurement in practical evaluation pipelines, where fully manual annotation is infeasible and judge-only metrics can be systematically biased.

\section{Background}
\label{sec:background}
\subsection{Problem Setup}
\label{sec:problem_setup}
We first formulate the problem of reference-based factuality evaluation for short-form query-response pairs, while noting the proposed methodology can be naturally extended to long form settings. Each evaluation item consists of a query $q_i$, a model response $a_i$, and the reference evidence $e_i$ obtained from retrieval, curated sources, or an existing evaluation pipeline. We write the observable input as $X_i=\{q_i,a_i,e_i\}$ and denote the human-annotated factuality label of each $X_i$ by $Y_i$. In practice, $Y$ is usually not a direct verdict, but verified atomic claims, or detected hallucinated spans.

In this paper, we focus on \emph{Factual Accuracy Rate} (FAR)\footnote{If the LLMaaJ’s evidence is the grounding provided to the answer generator, FAR measures faithfulness rate.}, which is defined as:
\begin{equation}
    \theta = \mathbb{E}[\phi(X,Y)],
\end{equation}
where $\phi(X_i,Y_i)$ denotes the factual correctness for the $i_{\text{th}}$ item.

\subsection{Judge-Assisted Factual Evaluation under Budgeted Annotation}
\label{sec:judge_assisted_estimation}

While FAR from human annotation is usually treated as the ground truth, acquiring human labels is expensive and slow, making it difficult to scale for drawing statistically significant conclusions. On the other hand, Automatic Factual Evaluation  can produce judge predictions at scale but the obtained FAR is generally biased. This motivates us to study a budgeted, judge-assisted factual evaluation problem where we combine the abundant judge predictions with limited human labels to form a statistically valid and more efficient estimator.

Given $N$ data points, an LLMaaJ provides predictions, denoted by $f(X)$, for all items, while only $n_{\text{bgt}} \ll N$ items can be sent for human annotation. We adopt Active Statistical Inference (ASI) \citep{asi}, which selects the labeled subset from an annotation policy, denoted by $\pi(X)$.

Let $\xi_i\sim \text{Bernoulli}(\pi_i)$ indicate whether item $i$ receives a human label with the inclusion probability $\pi_i$. ASI estimates FAR by:
\[
\hat{\theta}^{\text{ASI}}
=
\frac{1}{N}\sum_{i=1}^N
\left[
\phi_i^f+
\frac{\xi_i}{\pi_i}(\phi_i^h-\phi_i^f)
\right],
\]
where $\phi^f$ and $\phi^h$ denote the judge predicted and human annotated factual correctness respectively.

\begin{table*}[ht]
\setlength{\tabcolsep}{9pt}
\centering
\small
\begin{tabular}{p{0.30\linewidth}p{0.62\linewidth}}
\toprule
Signal family & Representative signals \\
\midrule
Judge-output artifacts (F1)
& Claim/span structure, verification or hallucination fraction, unverifiable outputs, self-reflection or confidence. \\

Evidence and reference quality (F2)
& Reference support strength, reference conflict, source freshness, direct vs. indirect support. \\

Task and input strata (F3)
& Task or domain type, query category, time-sensitive or location context indicators. \\

Answer-rubric alignment (F4)
& Claim importance, completeness vs. correctness, overprecision or auxiliary details, paraphrase/inference tolerance. \\
\bottomrule
\end{tabular}
\caption{Failure-space signal families for residual-risk policy design. The table lists representative pre-labeling signals rather than an exhaustive feature set.}
\label{tab:fsa_signal_families}
\end{table*}

\subsection{Statistical Efficiency}
\label{sec:statistical_efficiency}

Drawing statistically meaningful conclusions requires not only targeting the correct human-defined metric, but also estimating it with low variance under a limited annotation budget. We compare three estimators that are central to this paper. The \emph{classical} estimator, $\hat{\theta}^h = \frac{1}{n_{\text{bgt}}}\sum_{i:\xi_i=1}\phi_i^h$, uses only the $n_{\text{bgt}}$ human-labeled examples sampled uniformly at random. It is unbiased for the human-defined FAR, but can be noisy under small annotation budget. The \emph{judge-predicted} estimator, $\hat{\theta}^f = \frac{1}{N}\sum_{i=1}^N \phi_i^f$, uses judge predictions on all $N$ examples and therefore has low variance, but it is biased when the judge is systematically misaligned with human.

The ASI estimator remains human-anchored. However, its variance is policy dependent and scales with $\mathbb{E}[\Delta^2(1/\pi(X)-1)]$ , where $\Delta=\phi^h-\phi^f$ is the judge-human residual~\citep{asi}. Therefore, efficient annotation requires assigning higher sampling probabilities to examples with larger expected residuals, which motivates the failure-space analysis in the next section.

\section{FSA-guided Policy Design}
\label{sec:fsa_policy}

Efficient ASI requires allocating human labels to examples with large expected judge-human residuals. In the oracle case, the optimal policy is proportional to the conditional residual magnitude,
\[
    \pi^\star(X) \propto \sqrt{\mathbb{E}[\Delta^2 \mid X]}.
\]
However, since $\phi^h$ is unavailable before annotation, the practical policy-design problem is to predict residual risk from pre-labeling information: the query $q$, response $a$, reference evidence $e$, and judge output $f(X)$.

Directly modeling residual risk from raw text is difficult, and generic uncertainty is often insufficient for factuality evaluation. We therefore introduce \emph{failure-space analysis} (FSA), a factuality-specific procedure for deriving structured features $Z(X,f(X))$ that expose where judge-human misalignment is likely to occur. FSA organizes candidate policy signals into four families, summarized in Table~\ref{tab:fsa_signal_families}: \emph{judge-output artifacts}, \emph{evidence and reference quality}, \emph{task and input strata}, and \emph{answer-rubric alignment}. Together, these signals capture not only whether the judge is uncertain, but also whether the evidence is incomplete or stale, whether the query belongs to a difficult slice, and whether the response requires inference or rubric tolerance that the judge may handle differently from humans.

\begin{figure*}
    \centering
    \includegraphics[width=1.0\linewidth]{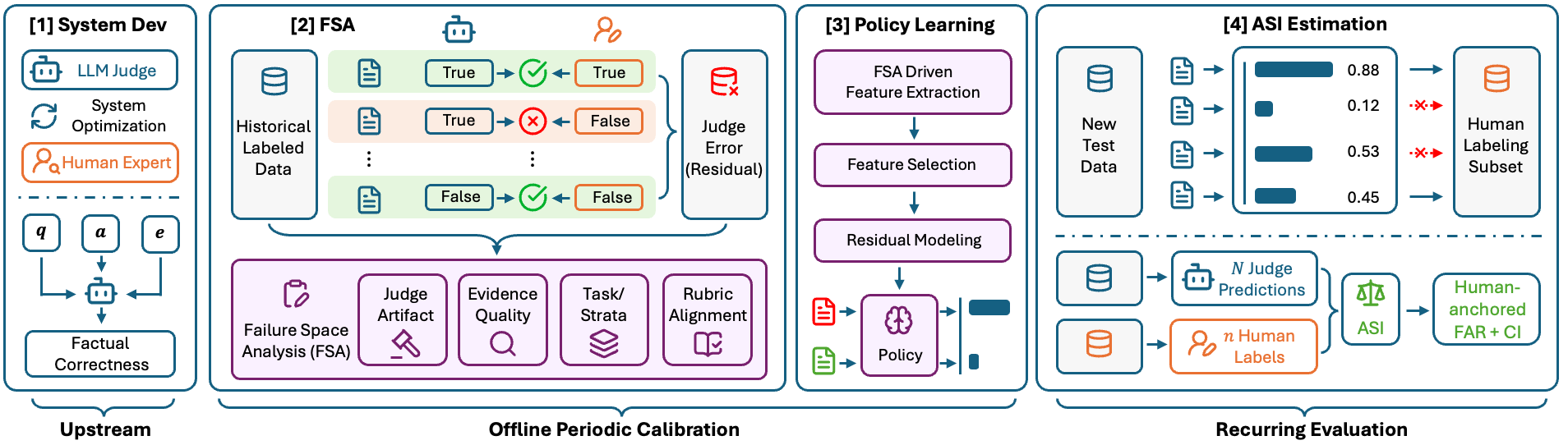}
    \caption{Overview of the pipeline of human anchored factual evaluation. [1] \textbf{System Development}: human experts design and iteratively optimize the LLM judge system, which takes the user query ($q$), model answer ($a$), retrieved evidence ($e$) and predicts the factual correctness. We consider this step as upstream and remains fixed once finished. [2] \textbf{Failure Space Analysis (FSA) on Calibration Data}: historical labeled data is leveraged to analyze the judge failure modes spreading across four main families. [3] \textbf{Policy Learning}: numerical features are extracted based on failure families and are selected with cross-validation. A regression model is trained to predict the risk of records based on the selected features, which is further converted to the sampling policy. High risk records are assigned higher probability to be sampled for human annotation. [4] \textbf{ASI estimation}: given new test dataset of size $N$, $n$ (labeling budget) of them are sampled with the policy and sent for human annotations, while all data receives judge predictions. The ASI estimator combines both judge predictions and human labels and produces human anchored factual accuracy rate (FAR) and confidence interval (CI).}
    \label{fig:overview}
\end{figure*}

Operationally, FSA uses a historical labeled set where both judge outputs and human annotations are available. We compute the residual target $|\Delta|$, extract numerical features from pre-labeling artifacts, and train a policy model to predict residual risk. The resulting scores are converted into sampling probabilities so that the annotation budget is concentrated on examples where the judge is most likely to disagree with humans. A pipeline overview is summarized and illustrated in Figure~\ref{fig:overview}. We then provide a comprehensive demonstration of this pipeline in Section~\ref{sec:disagreement_structure} and Section~\ref{sec:feature_modeling}.

\section{Experiments}
\label{sec:experiments}

\subsection{Evaluation Settings}
\label{sec:eval_settings}

We study two factuality-judge settings with different judge architectures, data formats, and judge output structures.

\paragraph{Automatic Factual Accuracy (AutoFA)}
AutoFA is an internal reference-based factuality evaluation system for voice assistant responses. Given a user query and model response, three reference answers are produced by search-enabled LLMs. The judge decomposes the response into verifiable claims, fact-checks each claim against the references, and produces an overall factuality verdict, cf. SAFE \cite{safe} and \textsc{VeriScore} \cite{veriscore}. Both claim-level and response-level outputs can be \textsc{true}, \textsc{false}, or \textsc{unknown}. We use $N=2338$ responses and FAR is defined as the fraction of responses rated factually correct among examples with a definitive verdict.

\paragraph{RAGTruth.}
RAGTruth~\citep{ragtruth} is a public hallucination benchmark with human span-level annotations. We use $N=2937$ GPT-4-0613 responses across question answering, summarization, and data-to-text generation. We evaluate Claude Sonnet 4.6 as a faithfulness judge using the benchmark prompt, which asks the judge to identify hallucinated spans. For RAGTruth, FAR is the faithfulness rate: the fraction of responses without hallucination.

\paragraph{Self-Confidence Score.}
Following prior work on confidence-driven inference~\citep{cdi}, we collect a self-reflected confidence score for each record by prompting the judge to assess its confidence in its own previous verdict, yielding a scalar uncertainty signal in $[0,1]$.

\subsection{FSA of Judge Residuals}
\label{sec:disagreement_structure}

We first compare judge FAR against human FAR to characterize the residuals that an annotation policy should target. On AutoFA, the judge underestimates human FAR by $13.3$ percentage points. On RAGTruth, the gap is larger: the judge FAR is $55.6\%$, compared with a human FAR of $86.2\%$.

\paragraph{Typical judge failures.}
In both datasets, judge-human disagreements are dominated by judge over-rejection. On AutoFA, $90\%$ of disagreements occur when the judge marks human-correct responses as \textsc{false} ($43\%$) or \textsc{unknown} ($46\%$). Manual inspection identifies two common mechanisms: \emph{evidence gaps}, where references do not directly state a reasonable answer that humans accept, and \emph{overly-strict verification}, where the judge rejects an otherwise correct response due to unsupported peripheral details. On RAGTruth, $94\%$ of disagreements occur when the judge flags hallucinations that humans do not annotate, mainly caused by \emph{overly-literal grounding}: the judge penalizes valid paraphrases or reasonable inferences because they are not explicitly stated in the reference.

\begin{table}[t]
\centering
\setlength{\tabcolsep}{4pt}
\small
\begin{tabular}{llccc}
\toprule
Dataset & Stratum & Prev. & $\mathbb{E}[\Delta^2]$ & Lift \\
\midrule
\multirow{2}{*}{AutoFA}
& Partial Verification & 0.32 & 0.41 & 2.58 \\
& Time-sensitive Query & 0.40 & 0.20 & 1.29 \\
\midrule
\multirow{2}{*}{RAGTruth}
& Low Self-confidence & 0.32 & 0.78 & 2.24 \\
& Judge flagged output & 0.43 & 0.70 & 2.02 \\
\bottomrule
\end{tabular}
\caption{Residual concentration across observable strata. Prevalence (Prev.) is the fraction of records in the stratum. Lift $>1$ indicates above-average residual density.}
\label{tab:disagreement_strata}
\end{table}

\paragraph{Non-uniform residual structure.}
Table~\ref{tab:disagreement_strata} shows that residuals concentrate in observable strata derived from judge outputs and task metadata. We report each stratum's prevalence, average squared residual $\mathbb{E}[\Delta^2]$, and \emph{lift}, defined as the ratio between stratum-level and population-level $\mathbb{E}[\Delta^2]$. On AutoFA, partial verification, i.e., cases where at least one extracted claim is marked \textsc{unknown}, has $2.58\times$ lift, while time-sensitive queries have $1.29\times$ lift. On RAGTruth, low self-confidence has $2.24\times$ lift, and judge-flagged outputs, i.e., cases where the judge returns one or more hallucination spans rather than an empty list, have $2.02\times$ lift. These concentrations show that judge-human residuals are structured rather than uniformly distributed, motivating the FSA-derived features in Section~\ref{sec:feature_modeling}.

\begin{table}[t]
\centering
\small
\setlength{\tabcolsep}{3pt}
\begin{tabular}{llcc}
\toprule
\multirow{2}{*}{Family} & \multirow{2}{*}{Feature} & \multicolumn{2}{c}{Spearman $|\rho|$} \\
 & & AutoFA & RAGTruth \\
\midrule
\multirow{2}{*}{Judge}
& self-confidence & $0.31$ & $\mathbf{0.66}$ \\
& flag-reliability & -- & $0.25^{\dagger}$ \\
\midrule
\multirow{2}{*}{Evidence}
& ref-support-strength & $\mathbf{0.57}$ & $0.22$ \\
& ref-temporal-consistency & $0.36$ & -- \\
\midrule
Strata
& time-sensitive / task & $0.32$ & $0.18$ \\
\midrule
\multirow{2}{*}{Rubric}
& inference-required & $0.44$ & $0.16$ \\
& qualitative-language & -- & $0.17$ \\
\bottomrule
\end{tabular}
\caption{Feature ranking by FSA family on absolute Spearman correlation ($|\rho|$) with $|\Delta|$. $^{\dagger}$Computed on the judge-flagged subset.}
\label{tab:feature_ranking}
\end{table}

\subsection{Residual-Risk Feature Modeling}
\label{sec:feature_modeling}

Guided by Section~\ref{sec:fsa_policy} and the residual patterns in Section~\ref{sec:disagreement_structure}, we instantiate dataset-specific features from the artifacts available before human annotation. For both datasets, we use the judge's self-confidence score as a generic uncertainty signal. 

For RAGTruth, where the judge may over-detect hallucinated spans, we additionally define \emph{flag-reliability} to assess whether the judge-flagged spans are plausibly unsupported by the reference.

For AutoFA, the dominant residuals are tied to reference quality, so we extract \emph{ref-support-strength}, which measures how directly the references support the response's core claims, and \emph{ref-temporal-consistency}, which measures whether search-generated references agree on time-dependent facts. We also include task-level strata, such as time-sensitive queries in AutoFA and task type in RAGTruth.

\begin{figure*}[t]
\centering
\includegraphics[width=\textwidth]{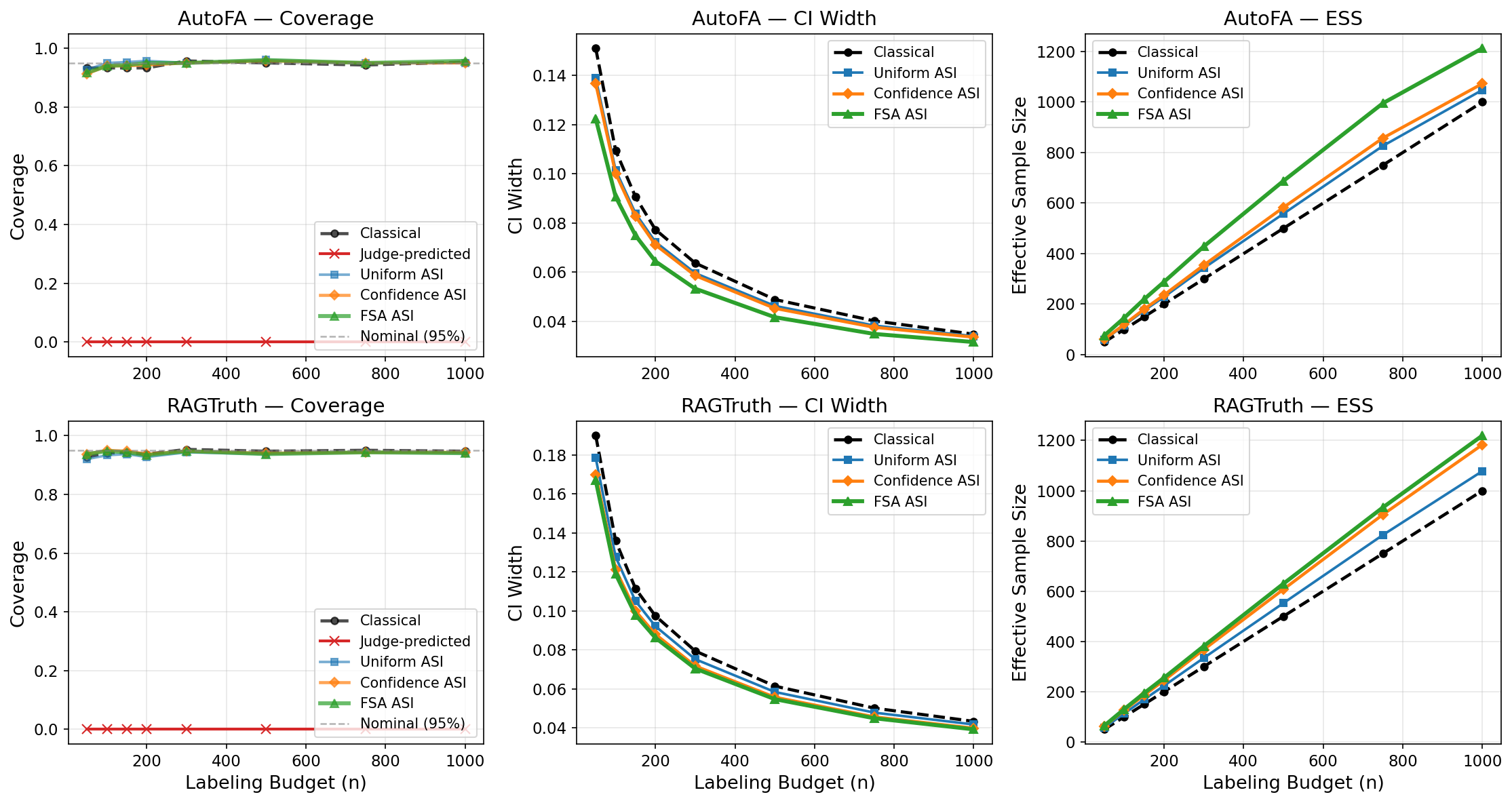}
\caption{Coverage, CI width, and effective sample size across annotation budgets. FSA ASI maintains valid coverage while achieving the largest effective sample size (ESS) across both datasets.}
\label{fig:ci_simulation}
\end{figure*}

Finally, we extract rubric-alignment features, including \emph{inference-required}, which measures whether verification requires reasoning beyond literal evidence, and \emph{qualitative-language}, which captures interpretive or subjective phrasing. LLM-extracted features are produced with structured prompts that return integer scores on a 0--3 scale. These feature-extraction calls are applied before human annotation and are substantially cheaper than human labeling in our setting. We provide more details in Appendix~\ref{feature_prompts}.

Table~\ref{tab:feature_ranking} reports the absolute Spearman correlation between each feature and the residual magnitude $|\Delta|$. This provides a univariate diagnostic of whether each feature monotonically tracks residual risk. On AutoFA, \emph{ref-support-strength} has the strongest correlation, consistent with evidence sufficiency being the dominant failure mode. On RAGTruth, \emph{self-confidence} dominates, while \emph{flag-reliability} provides an additional signal on the judge-flagged subset.

The correlations in Table~\ref{tab:feature_ranking} are diagnostic rather than the final policy criterion. A feature may correlate with $|\Delta|$ but still be less useful for sampling if it is sparse, redundant, or unstable under limited calibration data. We therefore split the calibration set into an inner training split for fitting residual-risk models and an inner validation split for comparing candidate feature configurations. This procedure selects \emph{ref-support-strength} for AutoFA, and \emph{self-confidence} together with \emph{flag-reliability} for RAGTruth, reflecting the different residual structures of the two factuality settings.

\subsection{Debiasing and Efficiency Evaluation}
\label{sec:policy_efficiency}

We evaluate the method along two dimensions: whether ASI recovers the human-anchored FAR despite judge bias, and whether FSA-derived policies improve annotation efficiency.

\paragraph{Policy learning.}
Each dataset is split 40/60 into calibration and evaluation sets. On the calibration set, we train a gradient-boosted tree (XGBoost) to predict $|\Delta|$ from the selected features, producing a residual-risk score for each record. The scores are converted into sampling probabilities by normalization with uniform mixing~\citep{robust_asi, asi}, and we apply power-tuning~\citep{ppi_pp} to set the judge contribution in the ASI estimator. Uniform-mixing and power-tuning parameters are estimated on the calibration set, with details in Appendix~\ref{app:policy_learning}. All reported results are evaluated on the evaluation split.

\paragraph{Estimators for comparison.}
We compare five estimators. \emph{Classical} uniformly samples $n_{\text{bgt}}$ human labels and estimates FAR with only human annotations. \emph{Judge-predicted} reports the judge FAR over all $N$ records without human correction. \emph{Uniform ASI} applies the power-tuned ASI correction under a uniform policy. \emph{Confidence ASI} uses only self-confidence as the sampling signal. \emph{FSA ASI} uses the FSA-selected features from Section~\ref{sec:feature_modeling}. The ASI variants differ only in how they allocate the human-labeling budget.

\begin{table}[t]
\centering
\small
\setlength{\tabcolsep}{4pt}
\begin{tabular}{llcc}
\toprule
Dataset & Policy & Cov. & ESS Gain (\%) \\
\midrule
\multirow{5}{*}{AutoFA}
& Classical     & $0.94{\scriptstyle\pm0.01}$ & --- \\
& Judge-predicted     & $0.00$ & --- \\
& Uniform ASI    & $0.95{\scriptstyle\pm0.01}$ & $+13.3{\scriptstyle\pm4.1}$ \\
& Confidence ASI & $0.94{\scriptstyle\pm0.01}$ & $+17.0{\scriptstyle\pm4.3}$ \\
& FSA ASI        & $0.95{\scriptstyle\pm0.01}$ & $\mathbf{+40.3}{\scriptstyle\pm9.1}$ \\
\midrule
\multirow{5}{*}{RAGTruth}
& Classical     & $0.94{\scriptstyle\pm0.01}$ & --- \\
& Judge-predicted     & $0.00$ & --- \\
& Uniform ASI    & $0.94{\scriptstyle\pm0.01}$ & $+11.3{\scriptstyle\pm1.8}$ \\
& Confidence ASI & $0.94{\scriptstyle\pm0.00}$ & $+22.6{\scriptstyle\pm2.3}$ \\
& FSA ASI        & $0.94{\scriptstyle\pm0.00}$ & $\mathbf{+27.1}{\scriptstyle\pm2.7}$ \\
\bottomrule
\end{tabular}
\caption{CI simulation results aggregated across budgets $n_{\mathrm{bgt}} \in \{50,100,\ldots,1000\}$ ($\alpha=0.05$, 300 trials per budget). Values report mean $\pm$ standard deviation across budgets.}
\label{tab:ci_results}
\end{table}

\paragraph{Metrics and simulation.}
We report coverage, confidence interval width, and effective sample size (ESS). Coverage is the fraction of confidence intervals containing the human FAR computed on the full evaluation set. ESS measures efficiency on the scale of uniform annotation: an ESS gain of $+40\%$ means that $n$ policy-sampled labels achieve the same variance as $1.4n$ uniformly sampled labels. For each budget $n_{\mathrm{bgt}} \in \{50,100,\ldots,1000\}$, we run 300 Monte Carlo trials on the evaluation split. In each trial, we bootstrap the evaluation set, sample human labels according to the corresponding policy, compute the estimator, and construct a Wald confidence interval at $\alpha=0.05$.

\paragraph{Results.}
Table~\ref{tab:ci_results} and Figure~\ref{fig:ci_simulation} summarize the simulation results. The judge-predicted estimator has zero coverage on both datasets, confirming that judge FAR is biased relative to the human-anchored target. In contrast, all ASI variants maintain near-nominal coverage across budgets, showing that human-labeled correction successfully debiases the judge estimate.

Among the valid estimators, all ASI variants improve efficiency over the classical baseline. Uniform ASI already provides ESS gains of $+13.3\%$ on AutoFA and $+11.3\%$ on RAGTruth while Confidence ASI further improves efficiency to $+17.0\%$ and $+22.6\%$, respectively, by allocating more labels to low-confidence examples.

FSA ASI achieves the largest gains, improving ESS by $+40.3\%$ on AutoFA and $+27.1\%$ on RAGTruth when aggregated across budgets. The larger gap over Confidence ASI on AutoFA reflects that self-confidence is a weaker residual-risk signal in this setting, while FSA identifies reference support strength as a more informative evidence-quality feature. On RAGTruth, the smaller gap is consistent with self-confidence already capturing much of the residual structure, with flag-reliability providing an additional gain.

\section{Related Work}
\label{sec:related_work}

Factuality evaluation is central to assessing LLM outputs when correctness must be verified against retrieved evidence, source documents, or external references. Prior work has developed automatic factuality and hallucination evaluation methods that decompose responses into claims, atomic facts, or hallucinated spans, and assess whether these units are supported by evidence~\citep{factscore, felm, ragtruth, openfactcheck, veriscore, safe}. These methods improve scalability, but LLM judges can systematically diverge from human labels due to incomplete evidence, over-literal grounding, rubric mismatch, and other biases in the judging procedure~\citep{mtbench, llmnotfair, surveyllmasajudge}.

Another line of works including Prediction-Powered Inference and its variants studies how to combine abundant machine predictions with limited human labels while preserving statistically valid inference~\citep{ppi, ppi_pp}. Active Statistical Inference extends this idea to budgeted annotation, where examples are sampled according to a policy designed to reduce estimator variance~\citep{asi}. Recent work adapts these ideas to LLM evaluation, showing how LLMaaJ metrics can be reported with valid uncertainty quantification~\citep{cdi, efficienct_eval_llm, how_to_correct}. Cost optimal AI evaluation~\citep{cost_optimal_eval} extends ASI by incorporating the cost of human annotator and LLMaaJ into the optimization objective to derive cost-aware annotation policy. Meanwhile, MultiPPI~\citep{multippi} and AM-PPI~\citep{amppi} explore settings where multiple predictors are available and how to route between predictors to balance the cost-performance trade-off.

Closest to our work, confidence-driven inference~\citep{cdi} uses verbalized LLM confidence to guide human annotation, which provides a strong generic sampling signal. However, factuality residuals are not always explained by uncertainty alone: judges can be confidently misaligned when evidence is incomplete or only indirectly supportive. Our work adds a factuality-specific policy-design layer: FSA derives residual-risk features from judge artifacts, evidence quality, task strata, and rubric alignment, improving annotation efficiency while preserving the human-anchored ASI target.

\section{Limitations}
\label{sec:limitations}
Our proposed framework is effective but is constrained by a few limitations: 

First, FSA-guided ASI relies on residual structure learned from historical or calibration data. Its efficiency depends on this structure remaining stable across the evaluation population, i.e., model updates, retrieval or grounding changes, shifts in user queries, or new failure modes can degrade the policy. Silent high-confidence errors that are not captured under distribution shift will be under-sampled by the policy, thus inflate inverse-probability weights and increase variance. Uniform mixing can mitigate but does not remove the need for distribution-shift monitoring and periodic policy recalibration in production deployment. Specifically, a small uniformly sampled audit stream can test whether recent LLM–human residuals remain concentrated in the regions predicted by the policy. If this relationship degrades, the system can increase uniform mixing or revert temporarily to Uniform ASI, while acquiring additional human annotations to recalibrate the policy.

Second, while the four failure families provide useful pre-annotation signals for predicting disagreement in the studied settings, they do not exhaustively cover all error categories. Rare but complex cases, for example, failures in multi-hop reasoning across multiple pieces of evidence, may still occur and may not be captured by FSA. Depending on the application setting, the FSA rubric may therefore need to be further adapted or extended.

Third, our formulation also simplifies the annotation process. We treat the final adjudicated human label as the target and do not explicitly model annotator-level noise or disagreement. We also focus on a single human-annotator rather than a multi-annotator setting with different costs and reliabilities, such as cheap annotators, expert annotators, or multiple judge models. Extending FSA-guided ASI to jointly allocate budget across heterogeneous annotators is an important direction for future work.

Finally, ASI improves efficiency for a given annotation budget, but it does not determine the budget required for a particular production decision. In practice, the choice of $n_{\text{bgt}}$ depends on cost, latency, acceptable uncertainty, and release-risk tolerance. Our results quantify the variance reduction achieved by FSA-guided sampling, while budget selection itself remains an operational decision.

\bibliography{reference}

\clearpage
% \onecolumn
\appendix

\section{Policy Learning Details}
\label{app:policy_learning}

This appendix provides experimental details on implementing FSA-policy and running simulations.

\begin{figure}[h]
\centering
\includegraphics[width=\linewidth]{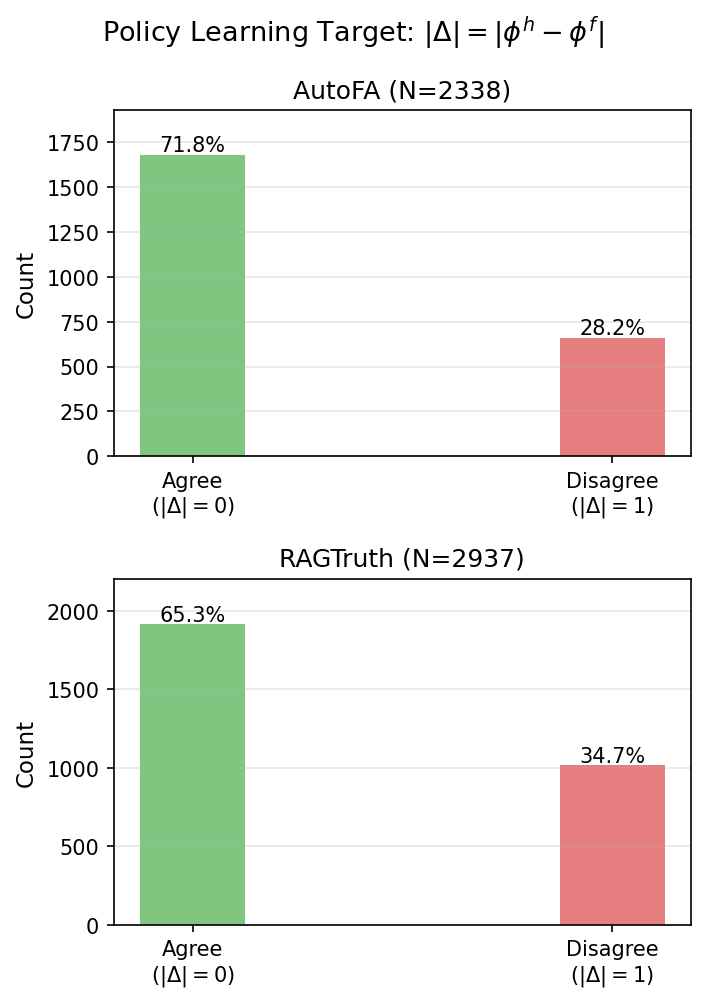}
\caption{Distribution of the policy learning target $|\Delta|=|\phi^h-\phi^f|$. Both datasets have an imbalanced binary target, with agreement (green) as the majority class.}
\label{fig:target_distribution}
\end{figure}

\paragraph{Regression target.}
The policy learning objective is to predict the residual magnitude $|\Delta_i| = |\phi^h_i - \phi^f_i|$ from pre-labeling features. On both datasets, this target is binary: $|\Delta|=1$ when the judge and human disagree, and $|\Delta|=0$ when they agree. Figure~\ref{fig:target_distribution} shows the class distribution. The disagreement rate is $28.2\%$ on AutoFA and $34.7\%$ on RAGTruth.

\paragraph{Score model.}
We train an XGBoost regressor~\citep{xgboost} with \texttt{objective=reg:squarederror} to predict $|\Delta|$ from the selected features. On a binary target, this is equivalent to learning $P(\Delta \neq 0 \mid X)$, producing a residual-risk score in $[0,1]$. We use 200 boosting rounds, learning rate 0.05, and maximum tree depth 3. The model is trained on the 40\% calibration split and its output serves as the raw score $s(X_i)$ for policy construction.

\paragraph{Score normalization and uniform mixing.}

Following the oracle ASI policy in Section 3, we first transform the predicted residual risk as \(r(X_i)=\sqrt{s(X_i)}\), and then convert it into inclusion probabilities:
\[
\pi_i^{\text{raw}} = r(X_i) \cdot \frac{n_{\text{bgt}}}{\sum_j r(X_j)},
\]
with water-filling to ensure $\pi_i \leq 1$. To prevent extreme weights that destabilize the ASI variance estimate, we also apply uniform mixing~\citep{robust_asi}:
\[
\pi_i = (1-\tau) \cdot \pi_i^{\text{raw}} + \tau \cdot \frac{n_{\text{bgt}}}{N},
\]
where $\tau \in [0,1]$ interpolates between the learned policy ($\tau=0$) and uniform sampling ($\tau=1$). Larger $\tau$ improves stability at the cost of policy concentration.

\begin{figure}[h]
\centering
\includegraphics[width=\linewidth]{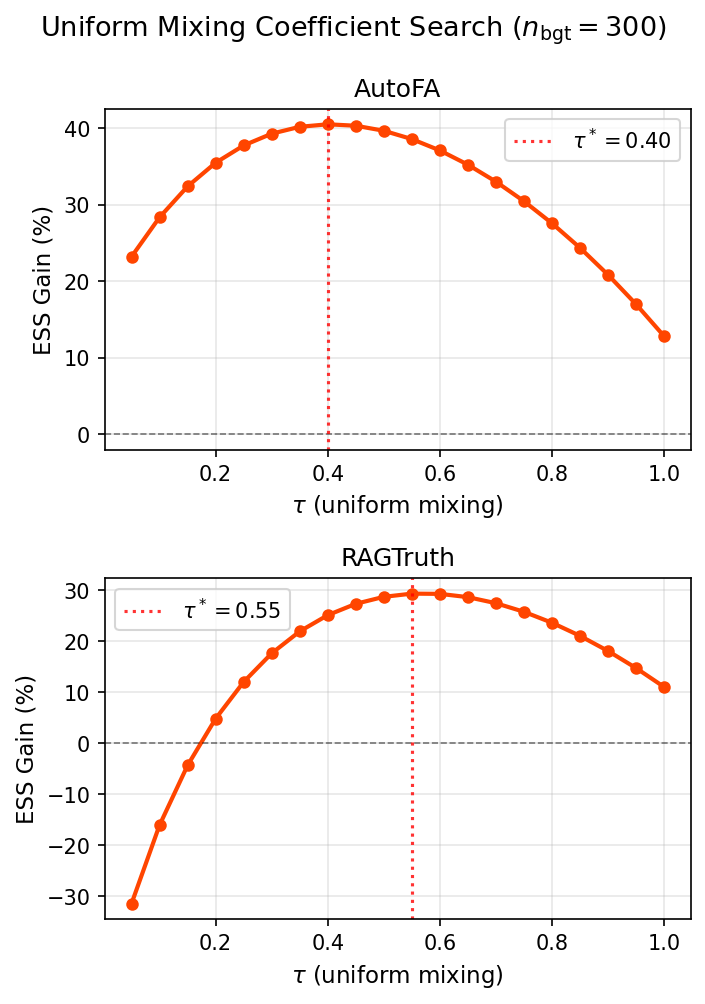}
\caption{Uniform mixing coefficient ($\tau$) search. ESS gain is evaluated on the calibration split at $n_{\text{bgt}}=300$. Both curves are concave: small $\tau$ overfits, large $\tau$ under-concentrates. Red dotted line marks the selected $\tau^*$.}
\label{fig:tau_search}
\end{figure}

\paragraph{Mixing coefficient selection.}
We select $\tau$ by grid search over $\{0.05, 0.10, \ldots, 1.0\}$, evaluating the ASI variance on the calibration split for each candidate. Figure~\ref{fig:tau_search} shows that both datasets exhibit a concave ESS-gain curve: too-small $\tau$ creates extreme inverse-probability weights that inflate variance, while too-large $\tau$ dilutes the policy signal. The selected values are $\tau=0.40$ for AutoFA and $\tau=0.55$ for RAGTruth. The higher $\tau$ on RAGTruth reflects its larger $\mathbb{E}[\Delta^2]/\mathrm{Var}(\phi^h)$ ratio, which requires a larger mixing coefficient for stability.

\paragraph{Power tuning.}
The ASI estimator uses a power-tuning coefficient $\lambda \in [0,1]$ that controls how strongly the judge contribution enters the correction~\citep{ppi_pp}. At $\lambda=1$, the estimator uses the full judge signal; at $\lambda=0$, it ignores the judge entirely. Power-tuning selects $\lambda$ to minimize the policy-dependent variance term:
\[
J(\lambda) = \mathbb{E}\!\left[(\phi^h - \lambda\phi^f)^2 \cdot \left(\frac{1}{\pi(X)} - 1\right)\right].
\]
Since $J(\lambda)$ is quadratic in $\lambda$, the optimum has a closed-form solution:
\[
\lambda^* = \frac{\mathbb{E}\!\left[\phi^h \phi^f \cdot w\right]}{\mathbb{E}\!\left[(\phi^f)^2 \cdot w\right]},
\]
where $w_i = 1/\pi(X_i) - 1$. In practice, we estimate $\lambda^*$ from the calibration split using sample averages. Because $J(\lambda)$ is convex, the tuned estimator always has variance no greater than the untuned ($\lambda=1$) estimator.

\begin{figure}[h]
\centering
\includegraphics[width=\linewidth]{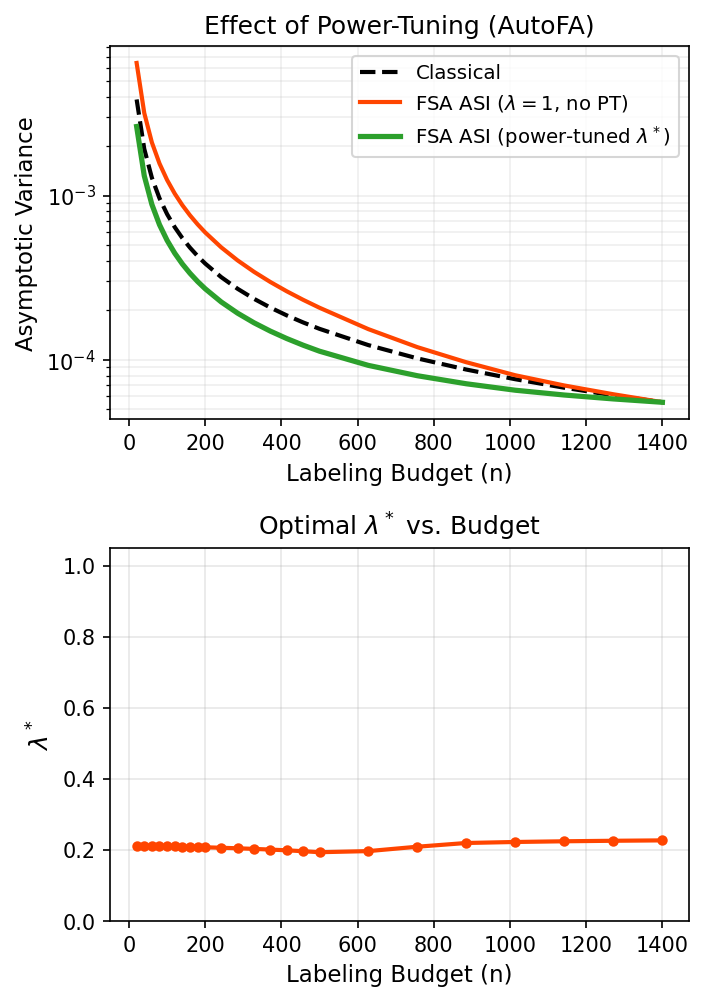}
\caption{Effect of power-tuning on AutoFA. Left: variance comparison showing that $\lambda=1$ (untuned) can be worse than baseline, while $\lambda^*$ (tuned) consistently improves. Right: the optimal $\lambda^*\approx 0.2$ is stable across budgets, indicating the judge should be used at $\sim$20\% strength.}
\label{fig:power_tuning}
\end{figure}

Figure~\ref{fig:power_tuning} illustrates the effect on AutoFA: without power-tuning ($\lambda=1$), the ASI variance exceeds the labeled baseline at small budgets because the judge residual is large. Power-tuning ($\lambda^* \approx 0.2$) shrinks the judge contribution to the level most appropriate for the residual structure, consistently reducing variance below the baseline.

\section{Feature Selection Details}
\label{feature_selection}

This appendix describes the feature selection procedure and provides post-hoc validation.

\paragraph{Selection procedure.}
Feature candidates are derived from the qualitative FSA in Section~\ref{sec:disagreement_structure}. We evaluate each candidate's residual-ranking quality via 5-fold cross-validated Spearman correlation on the calibration set. We then select the final feature set using two criteria: (1)~high CV Spearman, and (2)~parsimony: we prefer a smaller feature set because, with only $N \approx 1000$ calibration samples, using many features can introduce spurious correlations and cause the score model to overfit.

\begin{figure*}[h]
\centering
\includegraphics[width=0.49\linewidth]{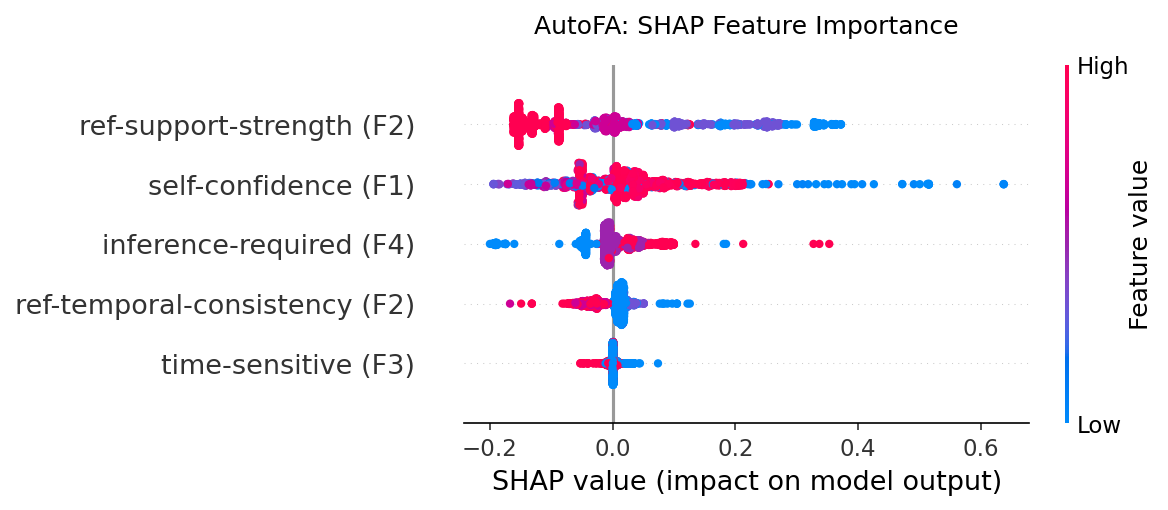}
\includegraphics[width=0.49\linewidth]{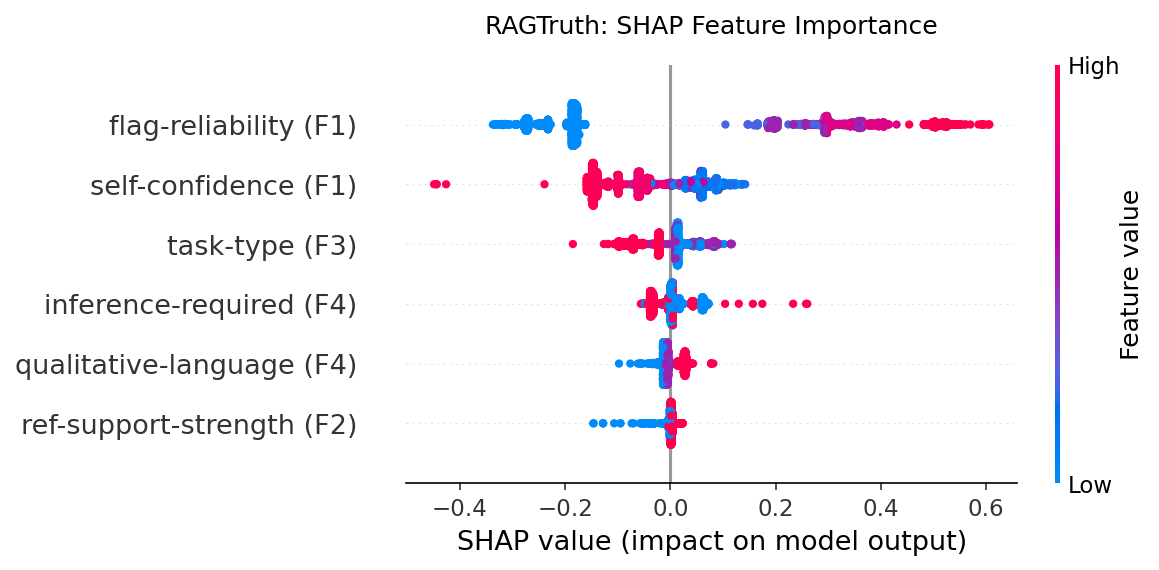}
\caption{SHAP feature importance for the residual-risk score model trained on all candidate features. Left: AutoFA. Right: RAGTruth.}
\label{fig:shap}
\end{figure*}

\paragraph{Joint feature importance.}
Figure~\ref{fig:shap} shows SHAP values~\citep{shap} for the score model trained on all candidate features, confirming which features drive the model's predictions. On AutoFA, ref-support-strength dominates. On RAGTruth, self-confidence and flag-reliability jointly dominate. Features from other families contribute marginally.

\paragraph{Post-hoc policy validation.}
To further evaluate how the selected features translate into effective policies, we compare held-out ESS gain across feature configurations in Table~\ref{tab:feature_selection_validation}. On AutoFA, ref-support-strength ($|\rho|=0.50$) alone outperforms self-confidence ($|\rho|=0.12$) alone by a large margin, validating that FSA identifies a stronger signal. On RAGTruth, self-confidence ($|\rho|=0.63$) is already strong; the addition of flag-reliability provides a modest improvement ($|\rho|=0.69$). In both cases, adding more features beyond the selected set does not improve ESS, consistent with the observation of overfitting at limited calibration size.

\begin{table}[h]
\centering
\small
\begin{tabular}{llcc}
\toprule
Dataset & Feature configuration & CV $\rho$ & ESS gain \\
\midrule
\multirow{3}{*}{AutoFA}
& Self-confidence only & 0.12 & $+16.6\%$ \\
& \textbf{Ref-support-strength} & \textbf{0.50} & $\mathbf{+40.5\%}$ \\
& All candidates (5) & 0.52 & $+24.7\%$ \\
\midrule
\multirow{3}{*}{RAGTruth}
& Self-confidence only & 0.63 & $+17.8\%$ \\
& \textbf{\makecell[l]{Self-confidence\\+ flag-reliability}} 
& \textbf{0.69} & $\mathbf{+19.7\%}$ \\
& All candidates (6) & 0.71 & $+9.6\%$ \\
\bottomrule
\end{tabular}

\caption{Feature selection validation. CV $\rho$: 5-fold cross-validated Spearman on calibration set. ESS gain: post-hoc evaluation on held-out split at $n_{\mathrm{bgt}}=300$. Bold: selected configuration. Adding more features improves CV correlation but can reduce ESS due to overfitting.}
\label{tab:feature_selection_validation}
\end{table}

The key observation is that higher univariate or multivariate correlation does not guarantee better policy performance. On AutoFA, the ``all features'' model achieves CV $\rho = 0.52$ but only $+24.7\%$ ESS, while the single-feature model achieves $+40.5\%$ ESS. This occurs because the multi-feature model requires more aggressive uniform mixing ($\tau = 0.70$) to stabilize extreme weights, which dilutes the policy concentration. The single feature provides smoother, full-population ranking that translates more effectively into sampling probabilities.

\section{Human Annotation Details}
Both datasets leverage human annotators to acquire reference labels and we summarize the details as follows: 
RAGTruth~\citep{ragtruth} reports two independent annotators per response and a third review for substantial disagreements, and consistency rates of $91.8\%$ at the response level and $78.8\%$ at the span level. Annotators were recruited through a professional vendor and compensated at $\$25$ per hour. 
For AutoFA, all records received a human annotation, and $90\%$ received a second, independent annotation. For records where two annotators disagree with each other on the final response factual verdict, a third human moderator was involved to adjudicate and made the final judgment. The human annotators are instructed to extract core, verifiable claims from the model responses and search the web to judge the factual correctness of the claims and final overall response level factual correctness.

\section{Qualitative Analysis}
We provide two qualitative examples from RAGTruth illustrating LLM judge errors and how FSA identifies them.

In record 168, the response states that Walter Scott was “shot whilst running from Officer Michael Slager.” The reference states that Scott was “fatally shot in the back by a police officer” and separately describes video showing Scott running from the officer while the officer fires eight shots. The response therefore combines two closely connected facts into a concise, well-supported statement, but the LLM judge flags “shot whilst running” as hallucinated, apparently requiring the temporal relation to be stated verbatim. Importantly, the LLM judge assigns 0.85 self-confidence, so confidence-based sampling would not identify this as a high-risk record. In contrast, the FSA feature flag-reliability receives a score of 2 (higher means LLM-judge is more likely to be wrong), indicating elevated over-detection risk and allowing the residual-risk policy to prioritize the record for human annotation.

In record 5718, the structured reference lists the same hours, 12:00–17:00, for every day from Monday through Sunday. The response accurately summarizes this as operating “seven days a week from 12:00 to 17:00,” yet the LLM judge flags the statement, apparently failing to aggregate the seven JSON entries. The LLM judge reports a low self-confidence score of 0.30 while the FSA assigns the maximum unreliability (flag-reliability=3).

\section{Feature Extraction Prompts}
\label{feature_prompts}

This appendix lists the prompts used for self-confidence scoring and FSA feature extraction. All prompts are written as Jinja2 templates: variables enclosed by \texttt{\{\{~\}\}} are replaced with record-specific content at inference time. The prompts are used only to extract auxiliary features or confidence scores, not to determine the final human label.

\subsection{Self-Confidence Prompt}

We use the prompt in Figure~\ref{fig:self-confidence-prompt} to elicit the judge's confidence in its own previous verdict. The same structure is used for both datasets, with minor wording changes depending on whether the underlying task is factuality verification or hallucination detection.

\begin{figure*}[t]
\centering
\begin{minipage}{\textwidth}
\begin{promptbox}{Self-Confidence Score: AutoFA Variant}
You previously evaluated whether a model response is factually correct by comparing it against reference answers, and concluded that the response is
\texttt{\{factually correct / factually incorrect\}}.

\medskip
\noindent\textbf{Inputs}

\begin{itemize}
    \item User query: \texttt{\{\{ query \}\}}
    \item Model response: \texttt{\{\{ response \}\}}
    \item Reference answer 1: \texttt{\{\{ ref\_1 \}\}}
    \item Reference answer 2: \texttt{\{\{ ref\_2 \}\}}
    \item Reference answer 3: \texttt{\{\{ ref\_3 \}\}}
    \item Previous verdict: \texttt{\{The response is factually correct / incorrect.\}}
\end{itemize}

\medskip
\noindent\textbf{Instruction}

How likely is it that your previous verdict is correct?

Output only a single number between 0 and 1 representing the probability. Do not provide an explanation or reasoning.

\medskip
\noindent\textbf{Output}

\texttt{Probability:}
\end{promptbox}
\end{minipage}
\caption{Prompt used for generating self-confidence score.}
\label{fig:self-confidence-prompt}
\end{figure*}

\subsection{AutoFA Feature Extraction Prompt}

The prompt in Figure~\ref{fig:autofa-prompt} extracts five FSA features for AutoFA. The judge assesses the verification difficulty of a response given the query and three reference answers. It is explicitly instructed not to make the final factuality decision.
\begin{figure*}[t]
\centering
\begin{minipage}{\textwidth}
\begin{promptbox}{AutoFA: Evidence and Rubric Features}
You are an experienced evaluator in factuality assessment for voice assistant responses.

Your job is not to decide the final label. Your job is to extract a compact set of interpretable features that characterize the difficulty of verifying whether the response is factually correct.

Use only the provided information. Return valid JSON only. Do not provide chain-of-thought. Use integer scores in \texttt{\{0,1,2,3\}}, where higher values indicate more of the corresponding feature.

\medskip
\noindent\textbf{Feature definitions}

\begin{enumerate}
    \item \texttt{reference\_temporal\_consistency}

    How consistent the three references are on time-dependent facts.

    \texttt{0} = fully consistent; \texttt{3} = major conflicts on the core answer.

    \item \texttt{reference\_support\_strength}

    How strongly the references support the response's core claims.

    \texttt{0} = no support; \texttt{3} = all references confirm the core answer.

    \item \texttt{inference\_required}

    How much inference is needed to verify the response against the references.

    \texttt{0} = pure extractive matching; \texttt{3} = substantial reasoning required.

    \item
    \texttt{claim\_importance\_variance}

    How much the claims vary in importance to the core question.

    \texttt{0} = all claims are equally important; \texttt{3} = core answer plus many peripheral claims.

    \item \texttt{answer\_completeness\_vs\_correctness}

    Whether the response appears correct but potentially incomplete.

    \texttt{0} = comprehensive; \texttt{3} = answers only a narrow slice.
\end{enumerate}

\medskip
\noindent\textbf{Output format}

\begin{verbatim}
{
  "feature_name": {
    "score": int,
    "justification": str
  },
  ...
}
\end{verbatim}

\medskip
\noindent\textbf{Inputs}

\begin{itemize}
    \item User query: \texttt{\{\{ query \}\}}
    \item Model response: \texttt{\{\{ response \}\}}
    \item Reference answer 1: \texttt{\{\{ ref\_1 \}\}}
    \item Reference answer 2: \texttt{\{\{ ref\_2 \}\}}
    \item Reference answer 3: \texttt{\{\{ ref\_3 \}\}}
\end{itemize}
\end{promptbox}
\end{minipage}
\caption{Prompt used for feature extraction for AutoFA.}
\label{fig:autofa-prompt}
\end{figure*}

\subsection{RAGTruth Feature Extraction Prompt}

The prompt in Figure~\ref{fig:ragtruth-prompt} extracts general verification-difficulty features for RAGTruth records. The general verification-difficulty features characterize the response–reference relationship, while \emph{flag\_reliability} additionally assesses the judge’s flagged spans.

\begin{figure*}[t]
\centering
\begin{minipage}{\textwidth}
\begin{promptbox}{RAGTruth: Evidence and Rubric Features}
You are an experienced evaluator in groundedness assessment.

Your task is to extract features that characterize the difficulty of verifying whether the response is faithful to the reference. If hallucination spans are provided, also assess whether those flagged spans are plausible.

Use only the provided information. Return valid JSON only. Do not provide chain-of-thought. Use integer scores in \texttt{\{0,1,2,3\}}, where higher values indicate more of the corresponding feature.

\medskip
\noindent\textbf{Feature definitions}
\begin{enumerate}
    \item \texttt{reference\_support\_strength}: how strongly the reference supports the response's claims. \texttt{0} = very weak support; \texttt{3} = strong and direct support.

    \item \texttt{inference\_required}: how much inference beyond verbatim extraction is needed to verify the response. \texttt{0} = purely extractive; \texttt{3} = substantial interpretation.

    \item \texttt{qualitative\_language}: degree of subjective or interpretive language in the response. \texttt{0} = purely factual; \texttt{3} = heavily qualitative.

    \item \texttt{flag\_reliability}: if hallucination spans are provided, how likely the judge is correct about those flags. \texttt{0} = very likely correct; \texttt{3} = likely incorrect due to over-detection. If no spans are provided, set this score to \texttt{0} and state that no flagged spans were provided.
\end{enumerate}

\medskip
\noindent\textbf{Output format}

Return a JSON object of the following form:

\smallskip
\noindent\texttt{\{\char`"feature\_name\char`": \{\char`"score\char`": int, \char`"justification\char`": str\}, ...\}}

\medskip
\noindent\textbf{Inputs}
\begin{itemize}
    \item Task type: \texttt{\{\{ task\_type \}\}}
    \item Reference: \texttt{\{\{ reference \}\}}
    \item Response: \texttt{\{\{ response \}\}}
    \item Judge's flagged spans: \texttt{\{\{ spans \}\}}
\end{itemize}
\end{promptbox}
\end{minipage}
\caption{Prompt used for feature extraction for RAGTruth}
\label{fig:ragtruth-prompt}
\end{figure*}

\end{document}